# Subsumptive reflection in SNOMED CT: a large description logic-based terminology for diagnosis

A.M Mohan Rao

mohan@ai-med.in

www.ai-med.in

68, Santosh Nagar, Mehidipatnam, Hyderabad-500 028, India

**Abstract.** Description logic (DL) based biomedical terminology (SNOMED CT) is used routinely in medical practice. However, diagnostic inference using such terminology is precluded by its complexity. Here we propose a model that simplifies these inferential components. We propose three concepts that classify clinical features and examined their effect on inference using SNOMED CT. We used PAIRS (Physician Assistant Artificial Intelligence Reference System) database (1964 findings for 485 disorders, 18 397 disease feature links) for our analysis. We also use a 50-million medical word corpus for estimating the vectors of disease-feature links. Our major results are 10% of finding-disorder links are concomitant in both assertion and negation where as 90% are either concomitant in assertion or negation. Logical implications of PAIRS data on SNOMED CT include 70% of the links do not share any common system while 18% share organ and 12% share both system and organ. Applications of these principles for inference are discussed and suggestions are made for deriving a diagnostic process using SNOMED CT. Limitations of these processes and suggestions for improvements are also discussed.

## 1. Introduction

Clinical decision support systems (CDSS) incorporate diagnostic decision and large amount of research has gone into developing suitable diagnostic algorithms. Examples of such systems include DxPlain[1], QMR-DT[2], PAIRS[3] and Isabel[4]. Typically a diagnostic engine runs on database to yield a diagnostic possibility. Predictive accuracy and hence assimilation into routine practice has been difficult for such products. Diagnostic process depends on weights for finding given disease, incidence of disease and statistical leak factors. However, weights remain main components in diagnosis which remains problematic since their pathophysiological processes are complex.

PAIRS data is indexed using SNOMED CT (September, 2015 version) for development of a Natural Language Processing (NLP) as well as Diagnostic Decision Support system (DDS) [3]. NLP involves keyword generation followed by search in relationship table[5]. Development of DDS involves application of probabilistic inference on weights of disease feature links. Bayesian probabilistic belief networks and its approximation method are developed for diagnostic decision [2, 6]. However, accuracy of these algorithms is based primarily on estimation of weights in database [6]. As databases are large and quantification of their pathophysiological processes are complex, development of DDS engines have become difficult.

Presently, automated inference under SNOMED CT is impossible for any reasoner for its complex hierarchies cannot be handled by them. Because of this as a recent review of SNOMED CT suggests, majority use it for design evaluations[7]. Therefore, it is of interest to identify simple inferential components that makes SNOMED CT adoption easier. The objective of present study is to propose a simple classification of clinical findings into 3 types:  a). concomitant in assertion and negation, b). concomitant in assertion alone, c). concomitant in negation alone. Further, logical implication (co-extension) of clinical finding is made in terms of a system and its organ. Extent to which a system and/or its organ are shared by disease- feature links is considered in estimation of their weights. For real world representation, the weights are normalized using phrase vectors of disease-feature links from a medical corpus[8] and disease incidence[9]. Finally a DDS is developed and hosted on web and its accuracy is tested and results are presented.

We use Is A relationship of SNOMED CT transitive closure table and classes for our analysis.  Each finding in PAIRS is assigned a unique index number that corresponds to SNOMED CT. An organ, system and pathophysiological process are also assigned for each by searching in the transitive closure table. PAIRS data includes 227 pathophysiological features, 58 organs for 12 major systems. It has logical implications to 12 root classes of SNOMED CT. Accuracy of DDS depends mainly on weights for features given disease. We use word2vec program[8] to derive vectors for findings in PAIRS data and these are used to improve its diagnostic performance. DDS engine is based on variational method as described by Jordan and Jaakkola[6]. Diagnostic accuracy is measured on NEJM clinical cases and their limitations are presented here.

## 2. Methods

### 2.1 Logical implication vs description logic in SNOMED CT

Logical implication or co-extension is an ontological concept used for truth evaluations[10]. It works in five steps: a). counter correlative, b) correlated substratum, c) adjunct limiting scope of counter correlative, d) adjunct limiting scope of correlated substratum and e) relation between adjuncts in a and d. In SNOMED CT context the steps would be: a). there is no myocardial infarction, b) chest pain, c) coronary artery disease, d) coronary angiogram, e) whenever there is coronary artery block there is myocardial infarction . These relationships can be traced across parent-child classes in SNOMED CT. Such a process, also known as logical implication helps in arriving at inference in description logic-based terminologies. Entities in a class share features of their parents. Since all classes inherit features of their parents and some parents have more than one child, one can reason that a parental class limits the scope of a child class. Since relationship between adjuncts limiting scope has bearing on inference, deriving a parental node becomes important. This logic as exemplified in the context of coronary artery disease can be generalized. These principles are used on PAIRS data for testing.

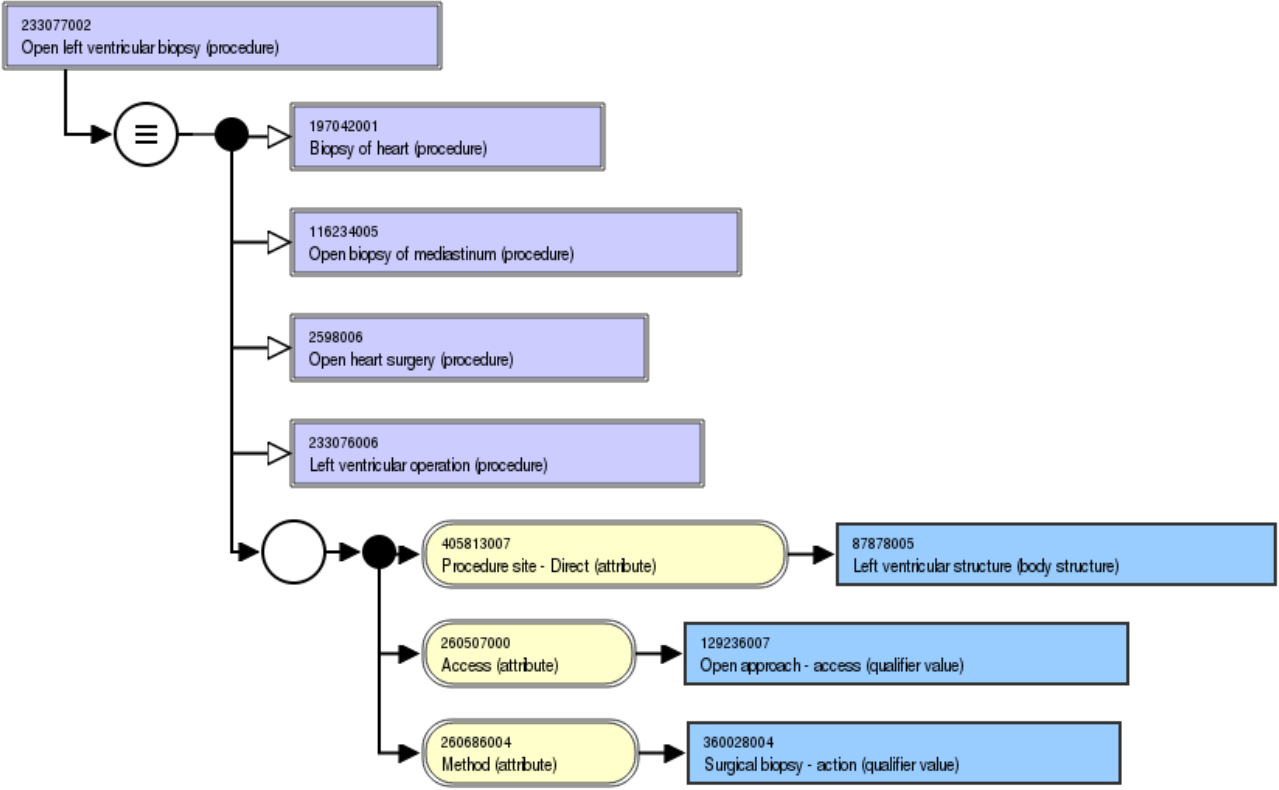

Fig.1 example shows a finding node (top left) is connected to organ (bottom right) of PAIRS data using SNOMED CT.

## 2.2 Dataset preparation

PAIRS disease feature links are collected from standard medical text sources. All diseases and features are identified using SNOMED CT unique ID. Features are classified as one of the categories: a) concomitant in assertion and negation, b) concomitant in assertion alone c) concomitant in negation alone. Transitive closure table is used to derive parents and their relationships and root classes are identified as: body structure, disorder, observable entity, finding, physical force, physical object, organism, procedure, product, situation, substance or value.

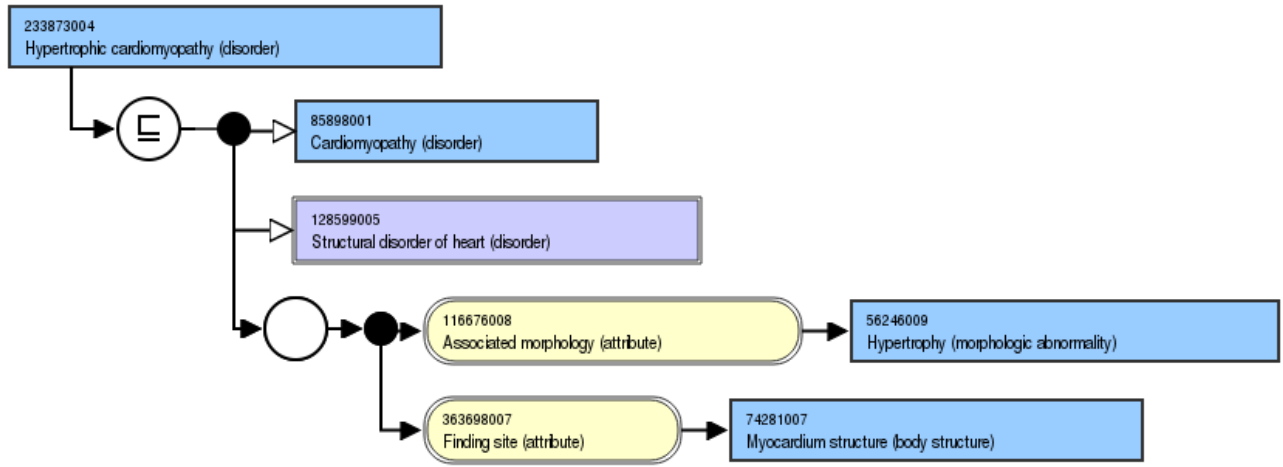

Fig. 2 example shows a disease node (top left) is connected to organ (bottom right) of PAIRS data using SNOMED CT.

**2.3 Finding of weights for features given disease**

Primary allocation of weights is based on concomitant assertion of features given disease. If presence of a feature confirms the disease and its absence negates the disease: it is defined as concomitant in both assertion and negation. For example, ventricular biopsy might confirm or negate cardiomyopathy as diagnosis. If a feature of a disease if present might suggest the disease but its absence might not rule out the disease: it is defined as concomitant only in assertion. Example includes dyspnea in cardiac failure. Similarly, absence of a feature might suggest a disease but its presence does not rule out that disease. Example includes absence of deep tendon reflexes in upper motor neuron lesions. Higher preference is given to those that are concomitant in both assertion and negation. Lower but equal preference is given to those that have only assertion or negation.

Condition of logical implication of a feature given disease shares common system or organ is considered for secondary allocation of weights. It is used for grouping them into: a) if both share common system, b) if both share different organ of same system, c) if both are of different system. Highest precedence is given for those that share common system compared to others. Similarly higher precedence is given for those that share different organs of same system compared to those that do not (see Fig 1 and 2).

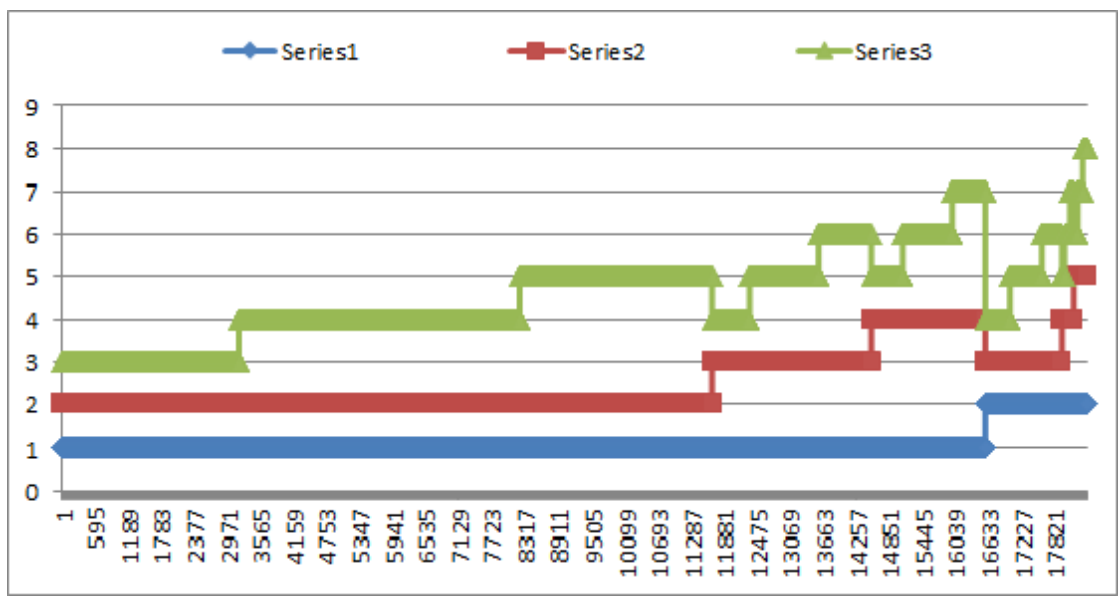

Fig. 3 Distribution of feature disease links in PAIRS data: concomitance in assertion or negation or both (series 1), shared system or organ (series 2) and inverse of vector distance (series3).

## 2.4 Vectors of features and diseases

Word2vec program is run on a 50 million word medical corpus for finding vectors for features and diseases. Each vector represents syntactic and semantic meaning of a word. We use these vectors for finding weights of features given disease. If a feature and given disease have small difference in their vectors, then they are closely related. In contrast if a feature and given disease have large difference in their vectors they are remotely connected. High preference is given for those that are closely related. Low preference is given to those that have large difference. If a feature given disease has intermediate difference in their vector they are preferred in between last two categories.

## 2.5 Allocation of weights to features given disease

The preferences described here are used to give weights to features given disease in 0.09 to 0.81 ranges. Estimation is done in 3 steps: a). features given disease preferences are sorted in tandem as described in 2.1, 2.2 and 2.3 (see Fig 3). b). those that have highest preference in all are allocated 0.81. Those that have highest preference by 2.1 & 2.2 but medium by 2.3 are given 0.72. Similarly those that have highest preference by 2.1 & 2.2 but low by 2.3 are considered 0.63. Similar allocations are made for ranges 0.36 to 0.54 and 0.09 to 0.27. The distribution of weights in PAIRS is as given in Fig 4. Large number of features given disease is in 0.09 to 0.27 range and few occupy 0.54 to 0.81 range. This suggests that there are fewer confirmatory findings for diseases.

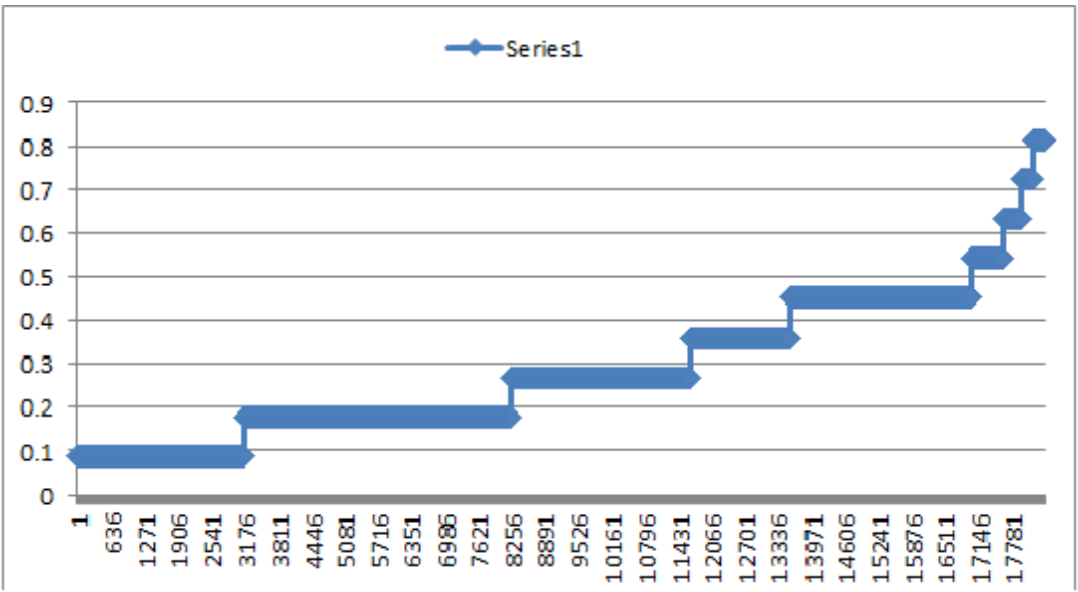

Fig. 4 Distribution of weights (0.09 to 0.81) for feature-disease links in PAIRS data.

**2.6 Diagnostic decision support engine**

DDS engine is based on variational method as described[6] and hosted on web at www.ai-med.in. A Natural Language Processor is developed using SNOMED CT algorithm and word tokens and their indices are stored in a database. Transitive closure tables are used to find relationships to features and their synonyms are identified from PAIRS database for display. The application takes either text or xml files for diagnosis. Diagnostic engine runs to generate lower bounds and a transformation process to generate diagnostic possibilities. Top 20 possibilities are displayed in form of infection, neoplasia, connective tissue disorder or others. It also suggests pathophysiological processes involved and tests needed for confirmation of diagnosis.

## 3. Results

**3.1 Quantitative analysis: weights for feature given disease in PAIRS**

The distribution of weights (0.09 to 0.81) in PAIRS data is given in Fig 4. The data is obtained by a 3 step process as described in methods (Fig 3). Large number of features (11 000) have weights in range of 0.09 to 0.36 where as a small number (1000) of them have weights in range of 0.63 to 0.81. This indicates that there is less number of confirmatory findings for diseases. Large number of features (16 569) are concomitant in assertion or negation. Small number of features (1828) is concomitant in assertion and negation (Fig 3). Logically features concomitant in assertion and negation are expected to be confirmatory, it corresponds to fewer number of them in 0.63 to 0.81 as stated and showed in Fig 4. Also large number in assertion or negation group indicates higher number of features in 0.09 to 0.36 ranges. Weights in range of 0.09 to 0.27 and 0.36 to 0.54 are each occupied by equal number of features (8 000). These two groups have majority features as expected in contrast to those in 0.63 to 0.81 ranges.

**3.2 Quantitative analysis: estimation of weights using SNOMED CT**

SNOMED CT class hierarchy is complex and its multiple inheritances and over use of Is A type relationships precludes reasoning by description logic[11]. We are able to allocate weights by introducing assertion and/or negation as a class for feature given disease. When used in conjunction with logical implication (or co-extension) of features given disease, they give a hierarchical preference for weights. As shown in Fig 3 large number of features given disease (13 031) do not share any system or organ while a small number of them (2288) share same system. Some features given disease (3078) share same system but not organ.

**3.3 Qualitative analysis: estimation of weights in PAIRS using SNOMED CT**

Performance of PAIRS features given disease is considerably improved by estimation of their weights since they are used in inference. PAIRS data uses these weights in its DDS. Thus description logic-based large terminologies SNOMED CT can be used to arrive at inference like diagnosis. As principles suggested here are simple these can be incorporated easily into existing architecture of any terminology.

**3.4 Evaluation of PAIRS**

We used MGH clinical cases of New England Journal of Medicine for evaluating PAIRS. We look for possibility of a diagnosis appearing in top 20 of PAIRS diagnoses. Actual diagnosis appears in top 20 of PAIRS diagnosis in 28 of 30 cases. The results are hosted at http://www.ai-med.in/cases/

Further diagnostic evaluations PAIRS are in progress.

## 4. Discussion

SNOMED CT is a comprehensive description logic-based terminology developed since 1965. It accumulates decades of knowledge of several thousands of doctors across the world. In spite of such enormous effort, there are short comings in terms of inference. By introduction of new inferential classes in it one can improve its performance. We used PAIRS as an example to show how one can use SNOMED CT for allocation of weights to features given disease. These weights are further used in arriving at a diagnosis in DDS.

Aristotelian logic uses 3 terms (major term, minor term and middle term) in deriving inference. Similar theory is proposed in Indian logic and an extensive coverage on perception and inference is available for over 5 millennia in Sanskrit literature[10]. Modern logic finds deficiencies in Aristotelean logic and several new theories on causation or inference are proposed. In context of description logic-based terminologies, reasoner (first order predicate logic based) like FACT++, pellet cannot cope with existing complexities of modern terminologies. Therefore, it is prudent to consider a different mechanism for inference which is eastern or Indian logic.

Indian logic based inference for diagnosis in SNOMED CT constitutes: a). Major term: diagnosis. b) Minor term: patient data other than clinical findings: age, sex, nationality. c) Middle term: clinical findings. A five step evaluation is made (see 2.1) followed by a co-extension of classes of features given disease. These principles are incorporated in PAIRS DDS engine to arrive at a diagnosis for given patient data. Performance of PAIRS is evaluated and results are available publicly.